\RequirePackage{fix-cm}
\documentclass[smallcondensed]{svjour3}     
\smartqed  
\usepackage{amsmath,amssymb,amsfonts}
\usepackage{graphicx}
\usepackage{textcomp}
\usepackage{xcolor}

\usepackage{subfigure}

\usepackage{diagbox}
\usepackage{threeparttable}
\usepackage{multirow}
\usepackage{color}
\usepackage{amsmath}
\usepackage{url}
\usepackage[misc]{ifsym}

\usepackage{booktabs}
\usepackage{caption}

\usepackage[ruled,linesnumbered]{algorithm2e}
\let\oldnl\nl
\newcommand{\nonl}{\renewcommand{\nl}{\let\nl\oldnl}} 

\begin{document}

\title{Online and Customizable Fairness-aware Learning
}


\author{Wenbin Zhang       
}

\authorrunning{Wenbin Zhang}


\institute{Wenbin Zhang\at Florida International University, Miami, FL 33199, USA  \\
	\email{wenbin.zhang@fiu.edu }           
}

\date{Received: date / Accepted: date}

\maketitle

\begin{abstract}
	While artificial intelligence (AI)-based decision-making systems are increasingly popular, significant concerns on the potential discrimination during the AI decision-making process have been observed. For example, the distribution of predictions is usually biased and dependents on the sensitive attributes (e.g., gender and ethnicity). Numerous approaches have therefore been proposed to develop decision-making systems that are discrimination-conscious by-design, which are typically batch-based and require the simultaneous availability of all the training data for model learning. However, in the real-world, the data streams usually come on the fly which requires the model to process each input data once ``on arrival'' and without the need for storage and reprocessing. In addition, the data streams might also evolve over time, which further requires the model to be able to simultaneously adapt to non-stationary data distributions and time-evolving bias patterns, with an effective and robust trade-off between accuracy and fairness. In this paper, we propose a novel framework of online decision tree with fairness in the data stream with possible distribution drifting. Specifically, first, we propose two novel fairness splitting criteria that encode the data as well as possible, while simultaneously removing dependence on the sensitive attributes, and further adapts to non-stationary distribution with fine-grained control when needed. Second, we propose two fairness decision tree online growth algorithms that fulfills different online fair decision-making requirements. Our experiments show that our algorithms are able to deal with discrimination in massive and non-stationary streaming environments, with a better trade-off between fairness and predictive performance.
\end{abstract}

\section{Introduction}
\label{sec:introduction}

Information systems are increasingly becoming automated and data-driven in both online and offline settings to render all sorts of decisions, including the allocation of resources, issuing of mortgage loans, assessment of pretrial risk, personalizing marketing and pre-screening and acceptance of applicants. However, as the usage of automated decision-making systems in assistance or even replacement of human-based decision making is widely adopted, growing concerns have been voiced about the potential loss of fairness and accountability in the employed models~\cite{datta2015automated,sweeney2013discrimination,zliobaite2015survey}. As a recent example, the AI algorithm behind Amazon Prime has exhibited signs of racial discrimination when deciding which areas of a city are eligible for advanced services~\cite{ingold2016amazon}. Areas densely populated by black people are excluded from services and amenities even though race is blind to the AI algorithm. Such incidents have sparked heated debate on the bias and discrimination in AI decision systems, pulling in scholars from a diverse of areas such as philosophy, law and public policy.


The basic aim of fairness-aware classifiers is then to make fair and accurate decisions, i.e., to train a decision-making model based on biased historical data such that it provides accurate predictions for future decision-making, yet does not discriminate deprived subgroups in the population. To this aim, a number of approaches have been proposed, ranging from discrimination discovery and elimination~\cite{hajian2016algorithmic,wang2020visual} to discrimination interpretation~\cite{aghaei2019learning,du2020fairness}, to ameliorate the intrinsic bias of the historical data~\cite{iosifidis2019fairness,kamiran2009classifying} in order to provide accurate and fairness-aware decision-making systems.

However, most of these studies tackle fairness as a static problem, assuming that the full data can be afforded for multiple scans and the characteristics of the underlying population do not evolve~\cite{hajian2016algorithmic,verma2018fairness,zafar2017fairness}. In many real-world applications though data is generated in a streaming fashion and its characteristics might also vary over time known as \emph{concept drift}~\cite{krawczyk2017ensemble}. Thus, it is necessary to design decision-making systems free of discrimination while considering the characteristics of online settings, which has been highly under-explored and brings unique challenges: \textbf{i) The massive data stream arrives continuously.} Discrimination-aware learning for such sort of applications therefore should be able to process each instance ``on arrival'' without the need for storage and reprocessing. However, most of the existing fairness approaches assume the full data can be scanned for multiple times, which is infeasible for such type of applications~\cite{iosifidis2019fairness}. \textbf{ii) The target concepts could also evolve.} Different from the static fairness methods assuming stationary data distribution by-design, in online settings the underlying data population might also evolve over time. Addressing both unfairness and evolving distribution simultaneously is challenging as the non-stationarity accompanies and complicates the biased decision regions. In addition, transferring approaches among these two domains is not straightforward, sophisticated design is thus warranted. \textbf{iii) Effective and robust trade-off control between fairness and accuracy.} Existing fairness-aware studies have largely been focused on maximizing fairness with the minimal loss in accuracy without an effective and robust mechanism to trade-off fairness and accuracy. However, when the class label is highly correlated with the sensitive attribute, enforcing such fairness constraints could result in a significant reduction in the predictive performance, thus results in underwhelming predictive performance and becomes inapplicable in terms of business objectives~\cite{zafar2019fairness}. In addition, such an effective and robust mechanism is also desired so as to allow for a customizable/fine-grained control of the level of fairness to instantiate application-wise fairness-aware learning. For instance, ``business necessity'' clause accounts for such scenarios stating some degree of discrimination could be relaxed to ensure fulfilling certain performance-related business needs, given such decision-making causes the least discrimination under the current performance-related constraints, i.e., accuracy~\cite{barocas2016big}.

To address the aforementioned challenges, this paper introduces a novel framework to design fair decision tree classifiers for online stream based decision-making. \emph{To the best of our knowledge, this is the first work capable of jointly addressing fairness, concept drift and customization.} We propose the new fair splitting criteria consider both data encoding and discrimination elimination, and additionally take evolving data distribution into consideration along with an effectively and robust trade-off between prediction accuracy and fairness performance. We then propose two any time fair decision tree growth algorithms to deal with discrimination in high-volume streaming environments, assuming stationary and non-stationary data distributions, respectively. The main contributions of this paper are:

\begin{itemize}
	\item We reformulate the information theory for the fairness-aware classification setting. Specifically, the fair information gain splitting criterion is introduced that jointly considers the information and the fairness gain of a split. We further introduce the adaptive fair information gain splitting criterion that additionally considers non-stationary distribution and fine-grained control of the trade-off between fairness and accuracy, thus providing more flexibility than state of the art approaches. 
	\item Two respective fairness-aware learners are designed for learning online from the massive stationary and non-stationary data streams that are common in many real-world applications. They inspect each instance in the stream on the fly for fair tree growing, which are able to achieve a good predictive performance over the stream while also preserving a low discrimination score. 
	\item Qualitative and quantitative experimental evaluations on a set of discriminated data streams demonstrate the utility of the proposed fairness-aware online learners in streaming environments.
\end{itemize}

The remainder of the paper is organized as follows. Section~\ref{sec: relatedWork} and~\ref{sec: background} review relevant work and the theoretical background knowledge regarding discrimination-aware learning, followed by the discussion of the base model, vanilla Hoeffding Tree, in Section~\ref{sec: ht}. Next, Section~\ref{sec: splittingCriteria} introduces the fair splitting criteria for stationary and non-stationary distributions, and the respective learning mechanisms for fairness-aware classification in online settings are presented in Section~\ref{sec: method}. Section~\ref{sec: experiment} then presents and analyzes the experimental results. Finally, we conclude the paper in Section~\ref{sec: conclusion}.


This work builds upon our prior research~\cite{zhang2019faht} and incorporates several noteworthy extensions, including: i) revising the fair splitting criterion previously introduced to accommodate zero discontinuity and negative multiplication impact, while still retaining its original desirable properties, ii) introducing a novel fair splitting criterion tailored to handle the massive streaming environment's non-stationary data distribution and enables fine-grained control over the trade-off between fairness and accuracy, iii) presenting a new algorithm for online fairness-aware tree growth in non-stationary settings based on the newly introduced fair splitting criterion, iv) enhancing the data preprocessing to better simulate distribution drift and enhance learning bias, v) conducting a comparative study of the newly proposed model against current state-of-the-art online fairness-aware methods.

\section{Related Work}
\label{sec: relatedWork}
Relevant to our work is work on fairness-aware learning as well as works on stream learning, in particular stream classification with decision tree models.

\subsection{Offline Fairness-aware Learning}
A number of research approaches have been proposed to address the problem of bias and discrimination in machine learning systems due to the inherent bias in data and the complex interaction between data and learning algorithms~\cite{pedreshi2008discrimination,zliobaite2015survey,hajian2016algorithmic}. 
These approaches generally fall into one of the following categories: i) pre-processing approaches, ii) in-processing approaches and iii) post-processing approaches.

\emph{Pre-processing approaches} modify the data distribution to ensure a fair representation of the different communities in the training set. The rationale for these approaches is that if a classifier is trained on discrimination-free data, its predictions will not be discriminatory (such a property however, cannot be guaranteed). 
A popular method in this category is massaging~\cite{kamiran2009classifying} that swaps the class labels of selected instances to restore balance. A ranker is used to carefully select the instances to be swapped in order to minimize the effect of label swapping on predictive accuracy; these are the instances closest to the decision boundary. 
Instead of intrusively relabeling the instances, reweighting~\cite{calders2009building} assigns different weights to different communities to alleviate the bias towards favored communities for the sake of benefiting the classification of deprived communities. The weights are assigned based on the expected and observed probability of an instance according to its sensitive attribute value and class. A higher weight will be assigned if the observed probability is lower than the expected probability to neutralize the discrimination. We emphasize that pre-processing approaches are typically not effective in eliminating discrimination arising from the learning algorithm itself. 

\emph{In-processing approaches} modify existing learning algorithms to also account for fairness instead of only predictive performance. Therefore, distinct from the first strategy which is classifier-agnostic, these approaches are algorithm-specific. For example, Aghaei et al.~\cite{aghaei2019learning} added regularization terms to the Mixed-Integer Programming model to penalize discrimination. Calders and Verwer~\cite{calders2010three} propose three approaches for fairness-aware Naïve Bayes classifiers: the first one alters the decision distribution until there is no more discrimination, the second one attempts to remove the correlation between sensitive attribute and class label by building a separate model for each sensitive group, and the third one models a latent variable to discover the actual class labels of a discrimination-free dataset. Closer to our work is the work by Kamiran et al.~\cite{kamiran2010discrimination} that incorporates discrimination into the splitting criterion of a decision tree classifier. Two key distinctions are: i) fairness in our work is directly defined in terms of the discrimination difference of the induction of a split, i.e., the fairness gain due to the split rather than the entropy w.r.t. sensitive attribute and ii) our model operates in an online setting rather than upon a static/batch dataset. The online learning scenario is much more challenging as decisions have to be made without access to the complete dataset and moreover, such decisions have an impact on following decisions in terms of both accuracy and fairness (feedback loops). Relevantly, the distance of an instance to the decision boundary is used in conjunction with the covariance between the sensitive feature and the decision boundary as a fair splitting criterion to enforce fairness in~\cite{nanfack2021boundary}. Such a splitting criterion is however not effective in the focused online setting of this paper as the decision boundaries/leaves are subject to change over time due to concept drift.


Finally, \emph{post-processing approaches} modify the resulting models by ``correcting'' the decision regions that lead to redlining for a fair representation of different subgroups in the final decision process. In~\cite{hajian2015discrimination}, the fair patterns are processed with $k$-anonymity to work against discrimination. Kamiran et al.~\cite{kamiran2010discrimination} carefully relabel selected leaves of a decision tree model to reduce discrimination with the least possible impact on accuracy. Transferring such approaches to a stream setting is not straightforward as the decision regions/leaves might change over the course of the stream due to concept drifts.

\subsection{Stream Classification}
The main challenge for learning in a stream environment is the so called, \emph{concept-drifts}, i.e., changes in the joint data distribution over time~\cite{krawczyk2017ensemble,aggarwal2007data,gama2014survey}. The learning methods therefore should be able to adapt to changes by learning incrementally from new instances, e.g.,~\cite{oza2005online,zhang2017hybrid} and by carefully considering historical information into the model by forgetting outdated information, e.g.,~\cite{read2019error,wagner2015ageing} and/or focusing on most recent data, e.g.,~\cite{bifet2007learning,zhang2017hybrid}. For example, the representative concept-adapting learner HAT that learns adaptively from data streams that drift over time with theoretical guarantees on performance~\cite{bifet2009adaptive}.



\subsection{Online Fairness-aware Learning}
\label{sec:onlineFairness}

A number of studies have been proposed with regard to offline fairness-aware learning and data stream learning solely focusing on the elimination of discrimination and concept drift, respectively. However, fairness in online setting requires simultaneously taking the removal of prediction dependence on the sensitive attributes and the evolution of underlying data distribution into consideration. In [18], massaging and reweighting are extended for online classification to introduce fairness enhancing interventions. Relevantly, an online fair boosting approach that changes the training distribution in an online fashion is proposed in~\cite{iosifidis2020online}. This model was later extended to update the distribution by also reducing the majority weights~\cite{iosifidis2021online}. In addition, online fairness approaches with individual constraints~\cite{gillen2018online}, taking the outputs and metadata of models as input~\cite{zhao2021fairness}, and human feedback~\cite{bechavod2019equal} have also been proposed. The critical limitation of these methods as well as other existing online fairness works is that they manage the concept drift implications on prediction accuracy and fairness performance separately as standalones. However, such implications could be simultaneous and further have an effect on each other. To alleviate such limitation, our approach jointly manages data encoding and bias reduction as well as adjusts the collectively learning focus on the fly depending on the respective non-stationary implications on prediction accuracy and fairness performance, i.e., the evolving relationships between the class and the features including sensitive attributes, thus be robust on the evolving implications of data encoding and bias reduction.

\section{Problem Definition}
\label{sec: background}
\sloppy
Assume a set of attributes $A= \{A_1,\cdots,A_d\}$ with their respective domains $dom(A_i)$ and let $Class$ be the class attribute of the classification problem. The data stream $D$ consists of massive and potentially infinite instances over the schema $(A_1,\cdots,A_d, Class)$ arriving over time, with each instance $x_t \in D$ being an element of $dom(A_1)\times \cdots dom(A_d)$ and class label $c_t \in dom(Class)$. Our learning setting is fully-supervised, however, the label of an instance $x_t$ arriving at time $t$ becomes available shortly after its arrival. Therefore, the goal is to make a prediction for $x_t$ at $t$ using
the current classifier and then, upon the arrival of its label to use
the labeled instance $(x_t, c_t), c_t \in dom(Class)$, for model update. This setup is known
as first-test-then-train or prequential evaluation~\cite{gama2010knowledge}.

We assume a special attribute $S$, referred as \emph{sensitive attribute} with a special value $s \in dom(S)$ referred as \emph{sensitive~value} that defines the discriminated community, i,e., the deprived community. Without the loss of generality, we assume that $S$ is a binary attribute: $dom(S)=\{s, \overline s \}$. As a running example, we use $S=$``gender'' as the sensitive attribute and $s=$``female'' as the sensitive value (with $\overline s=$``male''). We also assume the class is binary with values \{\emph{rejected}, \emph{granted}\}. By combining $S$ and \emph{Class} attributes (both binary), four communities are created:
\begin{itemize}
	\item \textbf{$s^{-}$} (deprived-rejected): females rejected a benefit.
	\item \textbf{$s^{+}$} (deprived-granted): females granted a benefit.
	\item \textbf{$\overline s^{-}$} (favored-rejected): males rejected a benefit.
	\item \textbf{$\overline s^{+}$} (favored-granted): males granted a benefit.
\end{itemize}

Formalizing fairness is a hard topic per se, there are more than twenty measures of fairness proposed in this direction but there is no consensus on which measure is more versatile than the others~\cite{verma2018fairness}. In this work, we adopt the widely used~\emph{statistical parity}~\cite{kamiran2009classifying} that examines whether the probability of being granted is the same for both deprived and favored communities.
Existing studies investigate statistical parity in static settings, we use its online variant, known as \emph{cumulative statistical parity}~\cite{iosifidis2019fairness}. More formally, for a set of instances $D$ from the given problem:

\begin{equation}
\label{equ: discrimination}
Disc(D_t)= \frac{\overline s^{+}_t}{\overline s^{-}_t+\overline s^{+}_t}- \frac{s_{t}^{+}}{s_{t}^{+} + s_{t}^{-}}
\end{equation}

\noindent where $D_t$ is the stream of $D$ at time $t$ while $s^{-}_t$, $s^{+}_t$, $\overline s^{-}_t$ and $\overline s^{+}_t$ are up to time $t$ the number of individuals from respective communities. If more of people of the deprived community are rejected a benefit comparing to the people of the favored community, the former could claim that they are discriminated up to time $t$. Such discriminative behavior can be evaluated in the original dataset $D$ or in the predictions of a classifier, capturing the fairness of the dataset representation or of the classifier, respectively.

Compared with fairness-aware learning in offline settings, addressing discrimination bias in online settings limits $D$ can only be scanned once and requires the classification task has a throughput at least equal to the rate at which each $x_t$ arrives. In addition, the concept drift could have implications on both data encoding and discrimination reduction, i.e., $P_{t_1}(D, c) \neq P_{t_2}(D, c)$ due to changes in the joint distribution for two different time-points $t_1$ and $t_2$ along with $Disc(D_t)$ evolves over time, the model therefore needs to adapt to such concept drift implications on the fly. Given the data stream $D$, the aim of online fairness-aware learning is then to maintain an up-to-date classifier $F$ which makes accurate predictions based on $A_i$ but also does not discriminate with respect to $S$ for infinite data streams.  

In this work, we acknowledge that the determination of what is considered ``fair'' or ``discriminatory'' is influenced by various factors and contextual considerations~\cite{Skirpan2017}, including philosophical inquiries that have been studied well before the interest of the AI community~\cite{binns2018fairness}. Although addressing such inquiries regarding the selection of statistical parity over other metrics, as well as more profound questions, is essential, it falls outside the scope of this paper. We have chosen statistical parity since it aligns with many users' intuition of ``fair'' decision-making according to American user studies~\cite{srivastava2019mathematical}. Hence, we anticipate that our approach will be useful in many applications, but it is necessary for informed users to carefully evaluate the relevant factors of their domain~\cite{Goldsmith2017}.

\section{Vanilla Hoeffding Tree}
\label{sec: ht}

Our online fairness-aware classifiers are built upon the Hoeffding Tree (HT) classifier~\cite{domingos2000mining}, one of the most popular data stream learners. The HT enjoys its popularity as it has sound guarantees of performance while ensuring a throughput, in which the tree built provably converges to that of a batch learner with sufficiently large instances.     

Given a stream of examples, the Hoeffding tree induction algorithm inspects each example in the stream only once and stores sufficient information in its leaves in order to grow during the induction of the tree. The sufficient information from the first examples will be used to choose the root test, then the succeeding examples will be sorted down into the corresponding leaves after the root attribute is chosen. This is also how prediction is being formed from a parallel procedure as needed at any point in time when processing training examples. The leaves store the sufficient statistics from the passed down examples for further growth and so on recursively. Compared with the batch decision tree classifier, the crucial decisions needed when constructing the HT are how but also when to split a node in handling with continuous flow of data. To this end, Domingos and Hulten employ the Hoeffding bound~\cite{maron1994hoeffding} to guarantee that a splitting decision is asymptotically nearly
identical to the decision of a conventional static learner. Such decisions aim to optimize for predictive performance and are based on \emph{information gain} (IG):

\begin{equation}
\label{equ: IG}
IG(D,A_i)= Entropy(D)-\sum_{v \in dom(A_i)}\frac{|D_v|}{|D|}Entropy(D_v)
\end{equation}

\noindent where $D_v, v \in dom(A_i)$ are the partitions induced by the attribute $A_i$ and the purity measurement \emph{entropy} of $D$ is:

\begin{equation}
\label{equ: entropy}
Entropy(D)= -\sum_{i} p(c_i) \log p(c_i)
\end{equation}

\noindent where $p(c_i)$ is the proportion of objects in class $c_i$.

A small entropy of $D$ implies small uncertainty, i.e., higher purity of $D$, and the uncertainty remaining in the entire dataset after splitting is defined by the weighted average entropy of each subset. The \emph{information gain} therefore looks for the splitting attribute that leads to the most reduction in entropy, i.e., the highest purity as the most objects in the partitions belong to the same class. Note that the value of IG falls within the range of [0,1].

As such, original HT aims to optimize for predictive performance and does not consider fairness. In addition, HT does not forget the old concept, adapting to concept drift by awaking the not growing leaves to restart growing when the under distribution is changing if evidence gathers that further splitting could better adapt to the current concept. However, such adaptation could be slow and therefore delayed or limited in practice.  

In this work, we extend the HT model in two ways to enable fairness-aware learning in massive and stationary settings as well as massive and concept-evolving scenarios. To this end, we first introduce two new splitting criteria: the \emph{fair information gain} that jointly considers the gain of an attribute split w.r.t. classification and also w.r.t. discrimination (Section~\ref{sec: FIG}), and the \emph{adaptive fair information gain} that additionally takes evolving data distribution and fine-grained control of fairness into consideration (Section~\ref{sec: AFIG}); we then propose two respective algorithms (Section~\ref{sec: FAHT} and~\ref{sec: 2CFAHT}) incorporating the newly proposed splitting criteria to fulfill different online fair decision-making requirements.

\section{The Fair Splitting Criteria}
\label{sec: splittingCriteria}

This section introduces two fairness-aware splitting criteria designed for stationary and non-stationary distributions, respectively. 

\subsection{The Fair Information Gain Splitting Criterion for Stationary Distribution}
\label{sec: FIG}

\begin{figure}[!t]
	\centering
	\makebox[\textwidth][l]{
		\subfigure[The HT induction]{
			\label{fig: toyExample_ht}
			\includegraphics[height=0.4\textheight,width=0.5\textwidth]{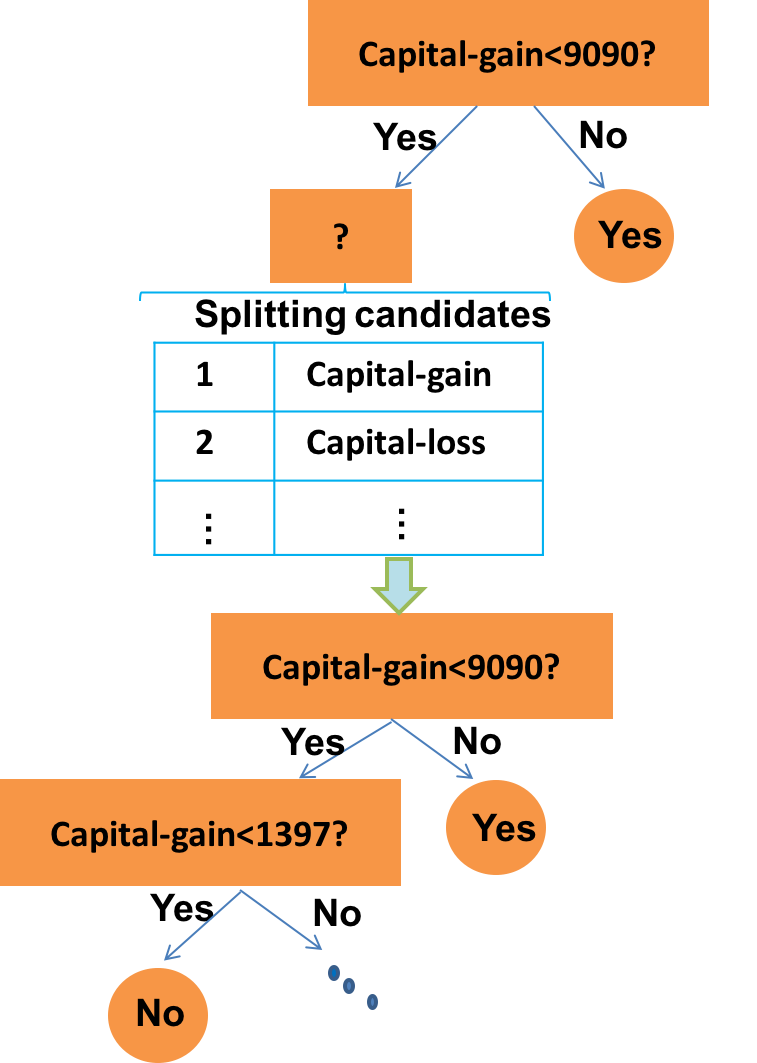}
		}%
		\subfigure[The FAHT induction]{
			\label{fig: toyExample_faht}
			\includegraphics[height=0.4\textheight,width=0.5\textwidth]{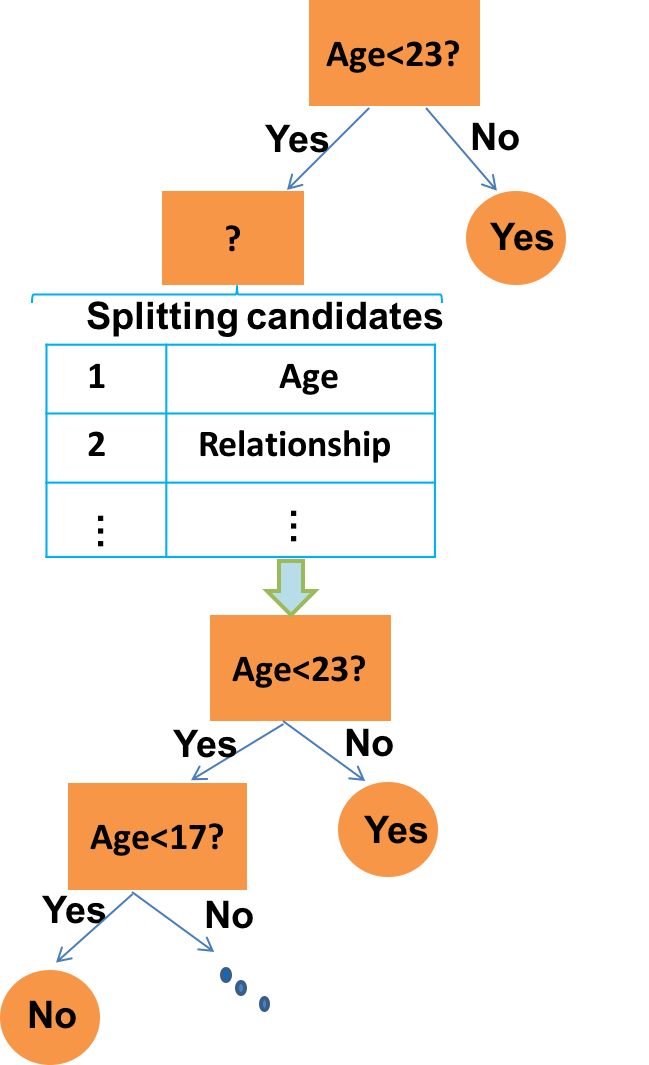}
		}%
	}
	\centering
	\caption{The comparison of attribute selection in tree inductions driven by different splitting criteria. Parts of the induced trees on the Adult dataset are shown; the sensitive feature and class label are ``gender'' and ``annual income'', respectively.}
	\label{fig: toyExample}
\end{figure}

The HT attribute-splitting decisions are exclusively accuracy-oriented and therefore an unfair tree might be induced over the course of the stream. Figure~\ref{fig: toyExample} illustrates parts of the trees induced from the accuracy-oriented HT as well as the proposed accuracy-driven and fairness-oriented FAHT (Section~\ref{sec: FAHT}) on the Adult dataset~\cite{Dua:2017}. As can be seen in Figure~\ref{fig: toyExample_ht}, the constructed HT selects ``capital-gain'' as the root node and re-selects at the succeeding nodes. This is quite intuitive as ``capital-gain'' is directly related with the class label ``annual income''. However, ``capital-gain'' also mirrors the intrinsic discrimination bias of the historic data as male, the favored community, has a higher likelihood of receiving higher capital-gain than female, the deprived community, based on the training data. Selecting such kind of attributes therefore naturally leads to a discriminated model so as to more encode the data for better accuracy. On the other hand, as shown in Figure~\ref{fig: toyExample_faht}, FAHT selects ``age'' as the root node and re-selects at the succeeding nodes as well. Compared with ``capital-gain'', selecting ``age'' as the splitting attribute also accounts for fair tree construction. The rational is that income per annum generally increases as one gets older and more experienced, which holds regardless of the sensitive attribute value. Selecting such an attribute therefore leads to an accurate and anti-discrimination model.

To balance data encoding and diminishing discrimination for an accuracy-driven as well as fairness-oriented model, we propose to alter the splitting criterion to also consider the fairness gain of a potential split. To this end, we first define the \emph{fairness gain} of an attribute $A_i$ relative to a collection of instances $D$ as the discrimination reduction in $D$ due to splitting on $A_i$:

\begin{equation}
\label{equ: FG}
FG(D,A_i)= |Disc(D)|-\sum_{v \in dom(A_i)}\frac{|D_v|}{|D|}|Disc(D_v)|
\end{equation}

\noindent where $D_v, v \in dom(A_i)$ are the partitions induced by $A_i$. In practice, the discrimination before splitting (i.e., on $D$) is compared against the discrimination after splitting, by aggregating the weighted discrimination on each resulting subset $D_v$. Its absolute value is used to capture reverse discrimination, i.e., discriminating against the favored community. The corresponding discrimination values for each partition are computed based on Equation~(\ref{equ: discrimination}). It should be noted that the value of FG is between -1 and 1 inclusive.

The idea of \emph{fairness gain} (FG) aligns with the idea of \emph{information gain} (IG) which measures the reduction in entropy resulting from a split; for both FG and IG it holds that the higher the reduction the merrier. The distinction is that the merit of FG is evaluated from the discrimination perspective while IG pays attention to accuracy. In addition, different from the previous work, FG is directly defined in terms of the reduction in discrimination due to a split rather than mediating between the entropy w.r.t. sensitive attribute as in~\cite{kamiran2010discrimination}. The limitation of such a mediation was acknowledged therein~\cite{kamiran2010discrimination}; in particular, their discrimination-aware splitting criteria do not lead to significant discrimination reduction unless being used in conjunction with additional leaf relabeling operations, thus combining in-with post-processing fairness interventions.

We combine \emph{fairness gain} and \emph{information gain} to a joint objective, called \emph{fair information gain} (FIG), that evaluates the suitability of a candidate splitting attribute in terms of both predictive performance and fairness. More formally, the \emph{fair information gain} of an attribute $A_i$ relative to a set of instances $D$ is defined as follows:


\begin{equation}
	\label{equ: ConjFairInfo}
	FIG(D,A_i) = (IG(D,A_i)+\xi) \times (FG(D,A_i)+1+\xi)
\end{equation}

\noindent where the constant 1 rescales the range of FG from [-1,1] to [0,2] to account for the negative impact of multiplication when combining FG and IG. Meanwhile, the small constant $\xi$ is also introduced to ensure the differentiability of the objective function, i.e., to prevent FIG from being zero.


Intuitively, the \emph{fair information gain} closely combines \emph{information gain} and \emph{fairness gain}. 
The iterative tree induction process therefore takes into account the influence of the splits under evaluation on the class distribution as well as the discrimination of the resulting tree, i.e., the construction of the tree is both accuracy- and fairness-driven. Note that the trade-off between accuracy and discrimination can be also achieved by combining IG and FG through other operations, e.g., addition. However, these two metrics are not necessary in the same scale in practice and the values of them can be incomparable, which means one can be dominated by the other. Multiplication is therefore favoured over addition. 

%

\subsection{The Adaptive Fair Information Gain Splitting Criterion for Non-stationary Distribution}
\label{sec: AFIG}

The previously introduced \emph{fair information gain} jointly considers the gain of an attribute split with respect to classification and also with respect to discrimination assuming the underlying distribution does not evolve over time. However, the incurred concept drifts when the data distribution is non-stationary could have implications on both accuracy and fairness. To adapt to such scenario, it is desired to adjust the jointly learning focus on the fly depending on the respective evolving implications on accuracy and fairness. To this end, we further propose the \emph{adaptive fair information gain} which adapts to the evolving relationships between the class and the features including sensitive attributes. Such flexibility also opens the door to adapt to applications with various performance constraints, thus enabling fairness-aware learning with fine-grained control.

In the massive data stream settings especially with non-stationarity, certain attribute values could discontinue for a period of time. Taking number and size of branches into account in FG when choosing an attribute therefore might not reflect the current attributes' representation. In addition, such representation favors global discrimination (i.e., attribute values dominate the majority of representations) over local discrimination (i.e., certain attribute values with a high discrimination rate but small in representation sizes), overweighting the majority groups more than the minority groups in avoiding discrimination. To effectively represent the discriminations in all groups evenly over the course of stream, we first reformulate the previous introduced FG and define the \emph{universal~fairness~gain}~(UFG) as follows,

\begin{equation}
\label{equ: UFG}
UFG(D,A_i)= |Disc(D)|-\sum_{v \in dom(A_i)}|Disc(D_v)|
\end{equation}

\noindent where all notations stay identical as in Equation~(\ref{equ: FG}).


In the general case, the more reduction in discrimination the better. When evaluating the fairness suitability of a candidate splitting attribute, a larger merit is therefore expected to be assigned if the attribute results in a higher discrimination reduction, regardless of the number of its distinct values and of each specific value. Compared with the previous definition, UFG removes the fairness gain ratio, which is equivalent to assigning a bonus to attributes with local discrimination but have a high discrimination reduction when being selected as the splitting attributes in Equation~(\ref{equ: FG}). That is to say, UFG encourages fair splits by giving priority to splitting candidates that result in a higher discrimination reduction regardless of their number of distinct values and less represented attribute values. The \emph{universal~fair~information~gain} (UFIG) can now be similarly defined in Equation~(\ref{equ: UFIG}) and to be used as the alternative fair-enhancing splitting criterion for the enhanced fairness-aware learning.  


\begin{equation}
\label{equ: UFIG}
UFIG(D,A_i) = (IG(D,A_i)+\xi) \times (FG(D,A_i)+1+\xi)
\end{equation}

According to the above definition, the attribute-splitting decisions are now accuracy-driven and fairness-oriented. However, it does not allow for a flexible control on the trade-off between fairness and accuracy. Neither does it account for the non-stationary underlying distribution. A clear mechanism is therefore expected to manage the trade-off in order to accommodate performance constrained applications as well as to cope with changes over time. In addition, when the accuracy and fairness come with different priority, we still want to always consider accuracy but the strength of the consideration on fairness is adjustable. To this end, we further reformulate UFIG by allowing adjustable weights of IG and UFG that contribute to the final determinacy while preserving other merits of UFIG. More formally, we define \emph{adaptive~fair~information~gain}~(AFIG) as:

\begin{equation}
\label{equ: AFIG}
AFIG(D,A_i, \gamma)= IG(D,A_i)\times e^{\gamma \times UFG(D,A_i)}
\end{equation}

In AFIG, the meaning of ``adaptive'' is two-fold. First, AFIG can adapt to applications with various performance constraints by presetting different $\gamma$ values which enforces respective weight/importance of fairness during the whole course of learning, thus allowing for fairness-aware learning with fine-grained control. Second, AFIG can also adapt to the evolving relationships between the class and the features including sensitive attributes by adjusting the value of $\gamma$ on the fly, thus be robust on the concept drift implications of data encoding and bias reduction. The on-the-go adapting mechanism of $\gamma$ will be detailed in Section~\ref{sec: 2CFAHT}.  

Compared to UFIG, AFIG also tightly combines UFG and IG except its exponential function along with an additional tunable parameter $\gamma$ to adjust the weight/importance of fairness, thus allowing for a fine-grained control of the level of fairness, and adapting to the evolving data distribution. When $\gamma$ is set as 0, i.e., accuracy is the primary focus of the current application, AFIG is identical to IG for the completely accuracy-driven model construction. For a positive $\gamma$ value, the merit of a feature increases with the discrimination reduction of that splitting feature and decreases with the resultant uncertainty increase. That is to say, the increment of $\gamma$ upweights UFG and correspondingly downweights IG. The model therefore favors features that result in a higher discrimination reduction for the more fairness-oriented model construction. Such flexible  design also opens the door for adaptive learning based on the respective implications on accuracy and fairness due to concept drift analogically. In other words, the model can adjust its learning focus according to the level of accuracy and fairness deteriorations due to the non-stationary data distribution. What's more, the exponential function in the formulation of AFIG is used for smoothing. For instance, suppose one attribute $A_a$ has a UFG of 0.1 and another attribute $A_b$ has a UFG of 0.01. Without the exponential function the weight of $A_a$ will be 10 times of that of $A_b$ assuming equal value of IG and $\gamma$ = 1. This may overly enforce fairness thus result in underwhelming accuracy. With the exponential function the weight for $A_a$ is 1.09 times of that of $A_b$.

While allowing for a fine-grained control of the trade-off between fairness and accuracy as well as adaptive learning based on the respective implications on these two metrics, AFIG still holds UFIG's other merits. That is to say, both of them are capable of working when IG and UFG are not in the same scale and encourage fair split by giving priority to splitting candidates that have positive UFG, i.e., splittings result in discrimination reduction. In addition, AFIG is identical to IG as well when UFG = 0, i.e., the splitting incurs no change in the level of discrimination.

\section{Online Fairness-aware Decision Tree Growth Algorithms}
\label{sec: method}

This section outlines two algorithms for online decision tree induction over stationary and non-stationary streams, as well as their respective specification of a number of refinements and modifications that instantiate the online fairness-aware learning.

\subsection{Tree Growth in Stationary Setting}
\label{sec: FAHT}

\subsubsection{FAHT: \underline{F}airness-\underline{A}ware \underline{H}oeffding \underline{T}ree Classifier}
\label{sec: alg_FAHT}
Our fairness-aware decision tree induction algorithm for stationary streams is built on top of the vanilla HT using the newly introduced \emph{fair information gain} splitting criterion (c.f., Section~\ref{sec: FIG}) that aims at optimizing both predictive performance and fairness of splits. This leads to the \emph{Fairness-Aware Hoeffding Tree} algorithm, shown in pseudo-code in Algorithm~\ref{alg: FAHT}.


\begin{algorithm}[!htb]
	\caption{FAHT induction algorithm}
	\label{alg: FAHT}
	\LinesNumbered
	\KwIn{a discriminated stationary data stream $D$,\\ ~~~~~~~~~~~confidence parameter $\delta$, tie breaking parameter $\tau$.}
	Let \emph{FAHT} be a tree with a single leaf (the root)\; 
	Init sufficient statistics at root\; 
	\For{each instance $x$ in $D$}{
		Sort instance into leaf $l$ using \emph{FAHT}\;
		Update sufficient statistics in $l$\;
		Increment $n_l$ counting the number of instances seen at $l$\;
		\If{examples seen at $l$ are not all of the same class}{
			Calculate fair splitting merit $\overline{G}_l(A_i)$ for each attribute\;
			Let $A_a$ be the attribute with highest $\overline{G}_l$\;
			Let $A_b$ be the attribute with second-highest $\overline{G}_l$\;
			Compute Hoeffding bound $\varepsilon= \sqrt{\dfrac{R^2\ln(1/\delta)}{2n_l}}$\;
			\If{$A_a \neq A_\emptyset$ and ($\overline{G}_l(A_a)- \overline{G}_l(A_b) > \epsilon$ or $\epsilon < \tau$)}{
				Split $l$ on $A_a$\;
				\For{each branch of the split}{
					Start a new leaf and initialize sufficient statistics\;
				}
			}
		}
	}
\end{algorithm}

FAHT starts out as a single root node (line 1) initializing the tree data structure (line 2), which stores sufficient statistics needed to compute the splitting merit for the induction of the tree. Each sequential instance from the $D$ is then sorted into an appropriate leaf $l$ using the tests present in FAHT built to that point (line 4), and updates $l$ including its sufficient statistics (line 5) and the instances count (line 6) accordingly. Lines 8-11 perform the splitting test related calculations, where $G$ is the splitting criterion function, e.g., the \emph{fair information gain}, with $\overline{G}$ as its estimated value. With these computed values, line 12 decides whether a particular attribute has won against the rest attributes including the null attribute $A_\emptyset$ according to the Hoeffding bound, of which R is the range of the splitting criterion. When the best splitting attribute has been selected, line 13-16 split the node, causing FAHT to grow.

\subsubsection{Details and Analysis of FAHT}
\label{sec: FAHT_system}

FAHT is built on top of the vanilla HT using the newly introduced \emph{fair information gain} splitting criterion (c.f., Section~\ref{sec: FIG}) for fairness-aware learning in massive and stationary online settings. Moreover, several refinements and modifications to the original HT are required to allow for a fairness-aware Hoeffding tree learner over discriminated data streams. These extensions are described hereafter. Note that other designs and theoretical guarantees still hold up and theorems, such as the induced tree is asymptotically nearly identical to the conventional static learner, can be proven accordingly. \\

\noindent
\textbf{Pre-pruning.} In the original HT algorithm, the idea of it might be more beneficial to not split a node at all is carried out by also considering the merit of no split, represented by the null attribute $A_\emptyset$ at each node for pre-pruning. Thus, a node is only allowed to split when the best split found is sufficiently better, according to the same Hoeffding bound test that determines differences between other attributes, than $A_\emptyset$. In our implementation, the merit to be maximized is the fair information gain. Therefore, the FIG of the best candidate split should be sufficiently better than that of the the non-splitting option. In terms of the FIG of the null attribute, the current class distribution is used to represent the IG and the FG is evaluated as the current level of discrimination.\\

\noindent
\textbf{Sufficient statistics.} The original HT algorithm scans each instance in the stream only once and stores sufficient information in its leaves to enable the calculation of the information gain for each possible split. In FAHT, additional information that necessitates the calculation of fair information gain should also be stored. 
In case of \emph{discrete attributes}, each node in the tree maintains a separate table per attribute. The counts of the class labels that apply for each attribute value are stored to calculate the information gain. In addition, the counts of each attribute value that keep track of the numbers of deprived and favored instances as well as of the instances receiving positive classification in deprived and favored groups are maintained to evaluate the fairness gain afforded by each possible split. The appropriate entries are updated incrementally based on new instances from the stream, based on attribute values, sensitive attribute values and class values. In case of \emph{numeric attributes}, each class label maintains its sufficient statistics as a separate Gaussian distribution. So are the four previous mentioned discrimination calculation related statistics. The updating of the numeric attribute involves updating appropriate distribution statistics according to the sensitive attribute values and class of the continuously arriving examples. The allowing test is used to select potential thresholds for binary splits and the merit of each allowed threshold is evaluated to select the most appropriate split among them. The merit is also computed according to the proposed fair information gain.\\

\noindent
\textbf{Memory.} 
The counts stored in the leaves are the sufficient statistics needed to compute the information gain afforded by each possible split. However, efficient storage is important. If there are $d$ attributes with a maximum number of $v=max_{A_i}|dom(A_i)|, 1 \leq i\leq d$ values per attribute and $c=|dom(Class)|$ possible classes in total, then a memory of $O(dvc)$ is required to store the necessary sufficient statistics at each leaf of the original HT algorithm. 
Our FAHT requires the maintenance of the four previous mentioned extra quantities in order to compute the proposed fair information gain measure. This is basically equivalent to add two more attributes, each with two values. Thus, the required memory becomes $O((d+2)vc)$ and therefore FAHT incurs negligible extra costs especially when $d \gg 2$. In the like manner, the continuous numeric attributes can also be stored efficiently for a slightly additional memory to compute the proposed heuristic measure afforded by each possible binary split.

\subsection{Tree Growth in Non-stationary Setting}
\label{sec: 2CFAHT}

FAHT is capable of learning from massive data streams, assuming that the distribution generating examples does not change over time. When the underlying distribution is non-stationary, FAHT does not forget the old concept, adapting to concept drift by awaking the not growing leaves to restart growing when the under distribution is changing if evidence gathers that further splitting could better adapt to the current concept. However, such adaption could be slow and therefore delayed or limited in practice. To this end, we further propose 2CFAHT which extends the FAHT model in two ways for massive non-stationary data streams: 1) by introducing an adaptive fair splitting criterion that adapts to the evolving concept drift implications on accuracy and fairness as well as applications with various performance constrains (i.e., Section~\ref{sec: AFIG}); 2) by adding the ability to detect and act more promptly to the evolution of underlying distribution and the incurred implications on accuracy and fairness (i.e., Section~\ref{sec: alg_2CFAHT}).

\subsubsection{2CFAHT: \underline{C}ustomizable and \underline{C}oncept-adapting \underline{F}airness-Aware \underline{H}oeffding \underline{T}ree Classifier}
\label{sec: alg_2CFAHT}

FAHT learns incrementally from the massive data streams by incorporating the incoming data in the stream into the model while simultaneously maintaining the performance of the classifier on the previous information. The tree is adapted based on the newly available data in the stream and does not forget the obsolete concept that does not follow the current example-generating process. While maintaining FAHT's speed, accuracy- and fairness-driven capabilities, 2CFAHT is also capable of adapting to evolving data distribution as well as managing the trade-off between fairness and accuracy by employing the previous introduced \emph{adaptive~fair~information~gain} as well as the ability of change detection and concept forgetting. To detect and react promptly to the evolution of the stream, 2CFAHT keeps its model consistent with the example-generating process of the current stream, creates and replaces alternative decision subtrees when evolving data distribution is detected at a node. Such drift can be reflected by whether there is a change in the error/fairness in that node. 2CFAHT extends FAHT which is incremental, so the tree is adapted by incorporating new instances into the model. Generally speaking, the performance of such model, under stationary distribution without drift, improves over the course of the stream as it generalizes better after incorporating more examples into the model. Therefore, performance deterioration is a good indicator of drift. 2CFAHT monitors the error/fairness rate of the non-leaf node to detect whether performance deteriorates and declare when branch replacement is necessary. However, evolving distribution may have different implications on error and fairness, for example deterioration in the accuracy while fairness increases. In addition, replacing branch according to one metric could have reverse impact on the other. To manage such trade-off on the fly, 2CFAHT leverages the previously introduced tunable parameter $\gamma$ to adjust the weight/importance of fairness according to current drift implications on accuracy and fairness, thus self-adapting to current distributions while managing the accuracy and fairness trade-off. The expression for such management is 

\begin{equation}
\label{equ: gamma}
\gamma_t = \left\{
\begin{array}{lr}
(1-a)\gamma_{t-1}, ~\mathrm{if}~ (b<0~\&~c \geq 0)~or~(b<c<0)\\
(1+a)\gamma_{t-1}, ~\mathrm{if}~ (c<0~\&~b \geq 0)~or~(b<c<0)\\
a\gamma_{t-1}, ~~~~~~~~~~~~~~~~\mathrm{otherwise} \\
\end{array}
\right.
\end{equation}

\noindent where $a$ ($0<a\leq1$) is the level of deterioration defined as follows:

\begin{equation}
\label{equ: a}
a = \left\{
\begin{array}{lr}
|b|, ~~~~~\mathrm{if}~b<0~\&~c \geq 0\\
|c|, ~~~~~\mathrm{if}~c<0~\&~b \geq 0\\
\frac{b-c}{c}, ~\mathrm{if}~b<c<0\\
\frac{c-b}{b}, ~\mathrm{if}~c<b<0\\
1, ~~~~~\mathrm{otherwise}
\end{array}
\right.
\end{equation}

\noindent where $b$ and $c$ are the degree of deterioration, i.e., the difference between the new performance and original performance divided by the original performance, due to concept drifts on accuracy and fairness, respectively. For $b$ and $c$, a negative value represents deterioration while non-negative values reflecting there is no negative concept drifts implication.

Specifically, when accuracy deteriorates but fairness does not, and the level of deterioration is $|b|$, the value of $\gamma$ correspondingly shrinks $(a \times 100)\%$ or $(|b| \times 100)\%$, that is to say $\gamma_{t}$= (1-$a$)$\gamma_{t-1}$, thus enforcing a higher accuracy weight during the current learning to compensate the negative impact on accuracy due to the current concept drifts; on the other hand, when fairness is negatively impacted by $(a \times 100)\%$ or $(|c| \times 100)\%$ but accuracy is not, the value of $\gamma$ increases by $a\%$, i.e., $\gamma_{t}$= (1+$a\%$)$\gamma_{t-1}$, to shift the current learning focus to fairness for the current negative concept drift implication on fairness; on the condition that accuracy and fairness deteriorate $|b|$ and $|c|$ respectively, the value of $a$ as well as the learning weight adjusting direction depend on the current concept drifts have more severe impacts on accuracy or fairness; otherwise accuracy and fairness do not decrease or decreased by the same level then the value of $\gamma$ remains, i.e., $a=1$. The use of such self-adapting strategy and sketch of 2CFAHT are shown in Algorithm~\ref{alg: 2CFAHT}.

\begin{algorithm}[h]
	\caption{2CFAHT induction algorithm}
	\label{alg: 2CFAHT}
	\LinesNumbered
	\KwIn{a discriminated non-stationary data stream $\widetilde D$,\\ ~~~~~~~~~~~confidence parameter $\delta$, tie breaking parameter $\tau$.}
	\nonl \textbf{ 2CFAHT($\widetilde D$, $\delta$, $\tau$)}\;  
	Let \emph{2CFAHT} be a tree with a single leaf (the root)\;
	Init sufficient statistics at root\;
	\For{each instance $x$ in $\widetilde D$}{
		Sort instance into leaf $l$ using \emph{2CFAHT}\;
		Update sufficient statistics in $l$ and nodes traversed in the sort\;
		Increment $n_l$ counting the number of instances seen at $l$\;
		\For{traversed node that has an alternate tree $T_{alt}$}{
			2CFAHT($x$, $\delta$, $\tau$)\;	
		}
		\If{examples seen at $l$ are not all of the same class}{
			Calculate adaptive fair splitting merit $\overline{G}_l(A_i)$ for each attribute\;
			Let $A_a$ be the attribute with highest $\overline{G}_l$\;
			Let $A_b$ be the attribute with second-highest $\overline{G}_l$\;
			Compute Hoeffding bound $\varepsilon= \sqrt{\dfrac{R^2\ln(1/\delta)}{2n_l}}$\;
			\If{$A_a \neq A_\emptyset$ and ($\overline{G}_l(A_a)- \overline{G}_l(A_b) > \epsilon$ or $\epsilon < \tau$)}{
				Split $l$ on $A_a$\;
				\For{each branch of the split}{
					Start a new leaf and initialize sufficient statistics\;
				}
			}
		}
		\For{non-leaf node that detects performance deterioration}{
			\uIf{$T_{alt}$== null}{
				Recalculating $\gamma_{alt}$ according to Equation~(\ref{equ: gamma})\;
				Create an alternative subtree $T_{alt}$\;
			}
			\Else{
				\uIf{$T_{alt}$ is more accurate or fair}{
					replace current node with its $T_{alt}$\;
				}
				\Else{
					prune its $T_{alt}$
				} 
			}
		}	
	}
	
\end{algorithm}

2CFAHT grows similarly to FAHT (line 1-2 and 10-21). The difference is that FAHT depends on fair splitting criterion while 2CFAHT employs adaptive fair splitting criterion to enable concept-adapting and fine-grained fairness-driven construction of the tree. What's more, in order to keep the model it is learning in sync with changes in the example-generating process, 2CFAHT continuously monitors the quality of old search decisions with respect to the latest instances from the data stream (line 22). 2CFAHT updates respective $\gamma$ to adjust learning focus accordingly (lines 24) then creates an alternative subtree (lines 25) for each node that change in the underlying distribution is detected. Under the condition that an alternative subtree already exists, 2CFAHT checks whether the alternative branch performs better than the old branch (line 27). The old branch will be replaced by the alternative one if so (line 28), otherwise the alternative branch will be pruned (line 30). Compared to FAHT, 2CFAHT also maintains sufficient statistics of the nodes traversed in the sort in order to update alternative branches (line 7-8). The learning process is therefore concept-adapting as well as allow for fine-grained control of fairness.

\subsubsection{Discussions and Further Extensions of 2CFAHT}

The proposed 2CFAHT induction algorithm extends the FAHT classifier, which is built on top of the vanilla Hoeffding Tree in order to extend its theoretical guarantees to the resulting algorithm while enabling online fairness-aware learning. 2CFAHT therefore still holds HT's theoretical guarantees and theorems, such as the induced tree is asymptotically nearly identical to the conventional static learner, can be proven accordingly. Moreover, 2CFAHT aims at adapting to evolving data distribution and fine-grained controlling fairness by further alleviating the discrimination bias towards the deprived group through the universal splitting evaluation and naturally exponentiating it as the fine-grained splitting criterion, the \emph{adaptive~fair~information~gain} (Section~\ref{sec: AFIG}), and by equipping itself with the ability of change detection and concept forgetting (Section~\ref{sec: alg_2CFAHT}). The modifications and refinements being included to Algorithm~\ref{alg: 2CFAHT} to instantiate the fine-grained fairness control and concept-adapting learning over non-stationary streams are similarity discussed in Section~\ref{sec: FAHT_system}. The differences are i) for pre-pruning, the AFIG of the best splitting candidate should be sufficiently better than that of the $A_\emptyset$ and $A_\emptyset$'s current level of discrimination is evaluated as UFG; ii) sufficient statistics should also be maintained and updated for alternative branches as well as the additional corresponding $\gamma$; and iii) although the required memory for each leaf or internal node is still $O((d+2)vc)$, 2CFAHT requires memory proportional to $O(n(d+2)vc)$ where is $n$ is the number of nodes in 2CFAHT's main tree and all alternate tress comparing to all leaves in FAHT. \\

\section{Experiments}
\label{sec: experiment}
\subsection{Experimental Goals and Evaluation Metrics}

The first goal of our experiments is to evaluate the prediction accuracy and fairness performance of our proposed FAHT and 2CFAHT methods, as is typical in the domain of fairness-aware machine learning~\cite{verma2018fairness}. To this end, we evaluate the different models in terms of accuracy and cumulative statistical parity. Due to the streaming nature of the data we use prequential evaluation, that is for each incoming instance from the stream we first predict its class via the model before updating the model~\cite{gama2010knowledge}. We report on both aggregated measures as well as on the over-the-stream performance of the different methods (Section~\ref{sec:exp_accu_fair}). Except for the landmark window model employed by the original HT, we also report on a sliding window model variation that focuses on the most recent history from the stream.

A second goal of our experiments is to understand the effects of our proposed fair information gain splitting criteria in the structure of the resulting decision tree models. To this end, we assess the impact of the newly proposed FIG and AFIG measures into the selected splitting attributes and eventually, in the tree structure as well as the latency in the splitting attribute selection process (Section~\ref{sec:exp_structure}). 

As one of the proposed decision tree learning algorithms is designed to process massive but also non-stationary data stream, the third goal of our experiments is to validate its concept adapting theoretical designs. To this end, we incorporate different concept adapting components into designs that do not consider concept drift to demonstrate the impact of non-stationary adaptations in the evolving data streams (Section~\ref{sec: justification}).

In addition, our learning model is designed to instantiate fine-grained control of the trade-off between fairness and accuracy for the sake of instantiating application-wise fairness-aware learning. Such fine-grained control is conveniently managed by adjusting the specifically introduced component, the tunable parameter $\gamma$ in our model. Section~\ref{sec:exp_structure} experimentally evaluates the effect of $\gamma$ in controlling fine-grained fairness-aware learning.

\subsection{Datasets}
Despite the growing concern on the discriminative behavior of AI models, there is still a lack of datasets and benchmarks that will allow for the development of new fairness-aware methods as well as for a systematic assessment of their capabilities~\cite{hajian2016algorithmic}. With respect to fairness-aware learning in data streams, this challenge is further magnified by the demanding requirement for big non-stationary datasets. 
Among the available datasets, the ones that best meet our requirements are the \emph{Adult}  and the \emph{Census} datasets~\cite{Dua:2017} both referring to the same learning task of predicting whether the annual income of a person will exceed 50K dollars. 
Other commonly used benchmarks such as the COMPAS and Default datasets~\cite{li2021time} are not applicable for online fairness evaluation as they are limited in the number of instances to simulate the streaming characteristics properly.

The \emph{Adult} dataset consists of 48,843 instances, each instance modeling a person in terms of 14 employment and demographic attributes. The sensitive feature is ``gender'' with female being the deprived group and male being the favored group, as in the previous studies~\cite{zafar2017fairness,zhang2019faht,zliobaite2015survey}. Making an annual income of more than 50K dollars is considered as receiving positive classification. The intrinsic discrimination level of the dataset is 19.45\%, based on Equation~(\ref{equ: discrimination}). The \emph{Census} dataset has the same learning task but is considerably larger with 299,285 instances and 41 attributes. The discrimination in the whole dataset is 7.63\% according to Equation~(\ref{equ: discrimination}). While existing works address these datasets from a static learning perspective, we render them as discriminated data streams by ordering the datasets based on the ``race'' attribute for the sake of better simulating concept and fairness drifts in the online setting, and then process them in sequence. 
The prequential evaluation~\cite{gama2010knowledge} is employed in which each incoming instance is first being predicted upon arrival and then is available for model training. 

%

\subsection{Accuracy vs. Fairness}
\label{sec:exp_accu_fair}

FAHT and 2CFAHT are designed to address the highly under-explored discrimination in massive data stream classification. To study the behavior of our approaches, we consider baselines from the following four perspectives: i) baseline model: the pure accuracy-driven \textbf{HT} which is also the base model of our methods to evaluate FAHT and 2CFAHT's anti-discrimination theoretical designs; ii) online fairness: the recently proposed fair online learner \textbf{FEI}~\cite{iosifidis2019fairness} and \textbf{FABBOO}~\cite{iosifidis2021online}, which are the only relevant online fairness work in the literature to the best of our knowledge (other competing online fairness
methods are not considered due to completely different setups as discussed in Section~\ref{sec:onlineFairness}); iii) fair splitting criterion: we incorporate into the HT model the static discrimination-aware splitting criterion of \cite{kamiran2010discrimination}, referred as \textbf{KHT}; iv) streaming method: the representative concept-adapting learner \textbf{HAT}~\cite{bifet2009adaptive}. The obtained results are summarized in Table~\ref{table: baselines}.

\begin{table}[!htb]
	\centering
	\caption{Accuracy-vs-discrimination between the proposed models and baseline methods. The best performance of the compared baselines is marked in boldface, so are the proposed models' when outperforming the respective performance of the best baseline method. Percentage in parenthesis is the relative difference over the performance of the best baseline method.}
	\label{table: baselines}
	\begin{tabular}{ccccc}
		\hline
		\multirow{2}{*}{\diagbox{Methods}{Metric}}& \multicolumn{2}{c}{\textbf{Adult dataset}} & \multicolumn{2}{c}{\textbf{Census dataset}} \\ 
		\cline{2-5}
		&  Disc  & Acc & Disc & Acc \\
		\hline
		HT & 24.14\% & 82.16\%  & \textbf{6.61\%} & 93.11\%\\
		FEI   &  \textbf{23.06\%}  & 74.27\% & 6.64\% & 80.06\% \\
		FABBOO   &  23.16\%  & 77.71\% & 7.32\% & 81.22\% \\
		KHT  &  24.24\% & 82.43\%  & 6.74\% & 93.26\% \\
		HAT   &  23.2\% & \textbf{84.62\%}  & 7.24\% & \textbf{94.06\%}\\
		\multirow{2}{*}{FAHT} &  \textbf{17.20\%} & 81.62\%  & \textbf{3.63\%} & 93.06\%\\
		&  (\textbf{-25.41\%})  & (\textbf{-3.55\%}) & (\textbf{-45.08\%}) & (\textbf{-1.06\%})   \\
		\multirow{2}{*}{2CFAHT}& \textbf{11.15\%} &  \textbf{85.01\%} & \textbf{0.88\%} & \textbf{95.03\%}\\ 
		&  (\textbf{-51.65\%})  & (\textbf{+0.46\%}) & (\textbf{-86.69\%}) & (\textbf{+1.03\%})   \\
		\hline
	\end{tabular}
\end{table}

In these results, it is clear that our models are capable of diminishing the discrimination to a lower level while maintaining a fairly comparable accuracy. Compared with the performance of the best baseline method, FAHT achieves a discrimination decrease of 25.41\% and 45.08\% at the cost of a slight 3.55\% and 1.06\% accuracy reduction on Adult and Census dataset, respectively, while 2CFAHT is able to further reduce discrimination by 51.65\% and 86.69\% with 0.46\% and 1.03\% accuracy enhancement. Due to the exclusively accuracy-oriented tree construction and the intrinsic discrimination bias of the historic data, a lack of fairness tree can be induced during the construction of the HT. Regarding KHT, little numerical difference is observed comparing to HT. The resulting trees of the methods look also pretty similar. Among all the baselines, HAT maintains the highest accuracy but also incurs a high discrimination. This is expected as HAT is exclusively accuracy-driven and ignores fairness. As also can be seen, although FEI and FABBOO are proposed for online fairness-aware learning, they perform poorly. This confirms our previous discussion that online fairness cannot be trivially solved by a simple combination of existing techniques from corresponding communities. We further claim that suit the remedy to the case theoretical designs of FAHT and 2CFAHT are fundamental to their reported outstanding performances, and to progress online fairness under respective stationary and non-stationary distributions rather than ad hoc.

From the previous results, it is clear that our methods seamlessly integrate the fairness merit and concept drift implications into the tree induction, results into models that are both accuracy-and fairness-driven as well as concept-adapting. FAHT is built upon HT and jointly considers fairness; such anti-discrimination capability of FAHT is also statistically significant as shown in Table~\ref{table: McNemarTest_faht}. 2CFAHT further considers the non-stationary nature of data streams when integrating fairness merit; such theoretical design can similarly be  statistically verified in Table~\ref{table: McNemarstest_2cfaht}.

%
%
\begin{table}
	\centering
	\caption{McNemar's test on deprived community between HT and FAHT applied to each dataset, which demonstrated FIG worked to benefit the positive classification of the deprived group in stationary distribution.}
	\label{table: McNemarTest_faht}
	\begin{threeparttable}[!htb]
		\begin{tabular}{ccccc}
			\hline
			\multirow{2}{*}{\diagbox{\textbf{HT}}{\textbf{FAHT}}}& \multicolumn{2}{c}{{\textbf{Adult dataset}}$^1$} & \multicolumn{2}{c}{{\textbf{Census dataset}}$^2$}\\ 
			\cline{2-5}
			& Granted &  Rejected & Granted & Rejected\\
			\hline
			Granted &  493 & 321 & 821 & 961 \\
			Rejected &  537 & 14,841 & 562 & 153,431 \\
			\hline
		\end{tabular}
		\begin{tablenotes} 
			\item[1] Chi-squared = 53.954, df = 1, p-value = 2.136e-13
			\item[2] Chi-squared = 104.01, df = 1, p-value $\textless$ 2.2e-16
		\end{tablenotes}
	\end{threeparttable}
\end{table}

\begin{table}
	\centering
	\caption{McNemar's test on deprived community between FAHT and 2CFAHT applied to each dataset, which demonstrated AFIG and evolution adapting design worked to benefit the positive classification of the deprived group in non-stationary distribution.}
	\label{table: McNemarstest_2cfaht}
	\begin{threeparttable}[!htb]
		\begin{tabular}{ccccc}
			\hline
			\multirow{2}{*}{\diagbox{\textbf{FAHT}}{\textbf{2CFAHT}}}& \multicolumn{2}{c}{{\textbf{Adult dataset}}$^1$} & \multicolumn{2}{c}{{\textbf{Census dataset}}$^2$}\\ 
			\cline{2-5}
			& Granted &  Rejected & Granted & Rejected\\
			\hline
			Granted &  617 & 413 & 861 & 522 \\
			Rejected &   546 & 14,616 & 640 & 153,752 \\
			\hline
		\end{tabular}
		\begin{tablenotes} 
			\item[1] Chi-squared = 18.169, df = 1, p-value = 2.022e-05
			\item[2] Chi-squared = 11.781, df = 1, p-value= 0.0005985
		\end{tablenotes}
	\end{threeparttable}
\end{table}

\begin{figure}[!tbh] 
	\centering
	\subfigure[Adult: accuracy]{
		\includegraphics[width=0.5\textwidth]{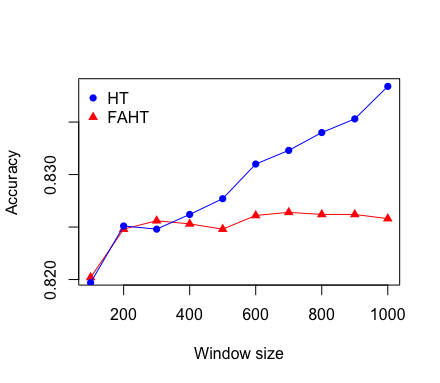}
	}%
	\subfigure[Adult: Discrimination]{
		\includegraphics[width=0.5\textwidth]{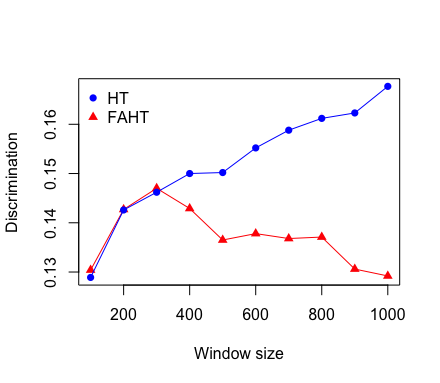}
	}
	\subfigure[Census: accuracy]{
		\includegraphics[width=0.5\textwidth]{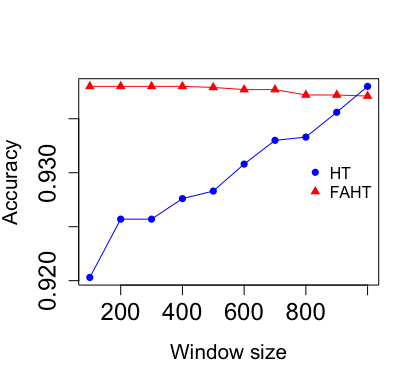}
	}%
	\subfigure[Census: Discrimination]{
		\includegraphics[width=0.5\textwidth]{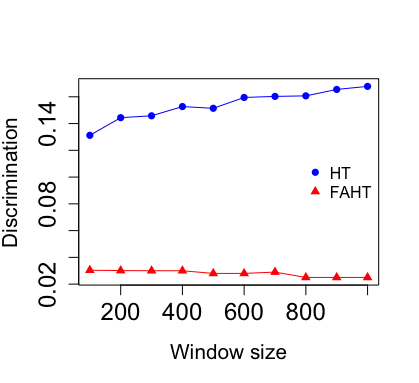}
	}%
	\centering
	\caption{The comparison of performance metrics in sliding window based data stream strategy. The data streams are processed in sliding windows; Each window trains a base learner as the ensemble component, the oldest one will be replaced when the classifier window is full; The ensemble members stored in the classifier window will also get updated with the instances in the current sliding window.}
	\label{fig: windowEnsembleResults_adult}
\end{figure}

Compared with 2CFAHT, FAHT is designed to process massive stationary data streams as it learns incrementally over the course of the stream and assumes that the distribution generating examples does not change over time. To evaluate the applicability of FAHT in evolving environments and how discrimination is propagated due to the stream processing, we employ FAHT as a base learner in sliding window based concept adapting strategy~\cite{zhang2017hybrid} to address concept drift. Specifically, the data stream is processed using sliding windows and a queue structured classifier window is maintained as the collection of learners from the previous streams. A new learner is trained based on the most recent sliding window and will be added to contribute for the ensemble learning. Again, the HT and FAHT are employed as the based learners for comparison. During the learning process, the components of the ensemble classifier are trained incrementally with the instances in the new sliding window. The obtained results are depicted in Figure~\ref{fig: windowEnsembleResults_adult}. 

As one can see, FAHT consistently pushes the discrimination to a lower level while maintaining a fairly comparable accuracy when being used as the base learner in the sliding window based concept adapting strategy. In particular, FAHT achieves the best discrimination reduction while HT gives the worst fairness result when the window size is 1000 on both datasets. This is not surprising; with the updates discrimination inherited from the previous stream could propagate and persist in later predictions. Therefore, although HT provides a better prediction performance, it leads to an unfair  model. On the other hand, the fair information gain splitting criterion of FAHT helps to control the discrimination propagation and manages to push the discrimination to a low level while maintaining a high prediction capability. These results are consistent with the results from the previous incremental-learning fashion and show the discrimination-aware learning ability of the proposed method. Therefore, the applicability and anti-discrimination propagation merit of FAHT hold.


\subsection{Structural Effects on the Tree Construction}
\label{sec:exp_structure}
\begin{figure}[!t]
	\centering
	\includegraphics[width=0.8\textwidth, height=0.5\textheight]{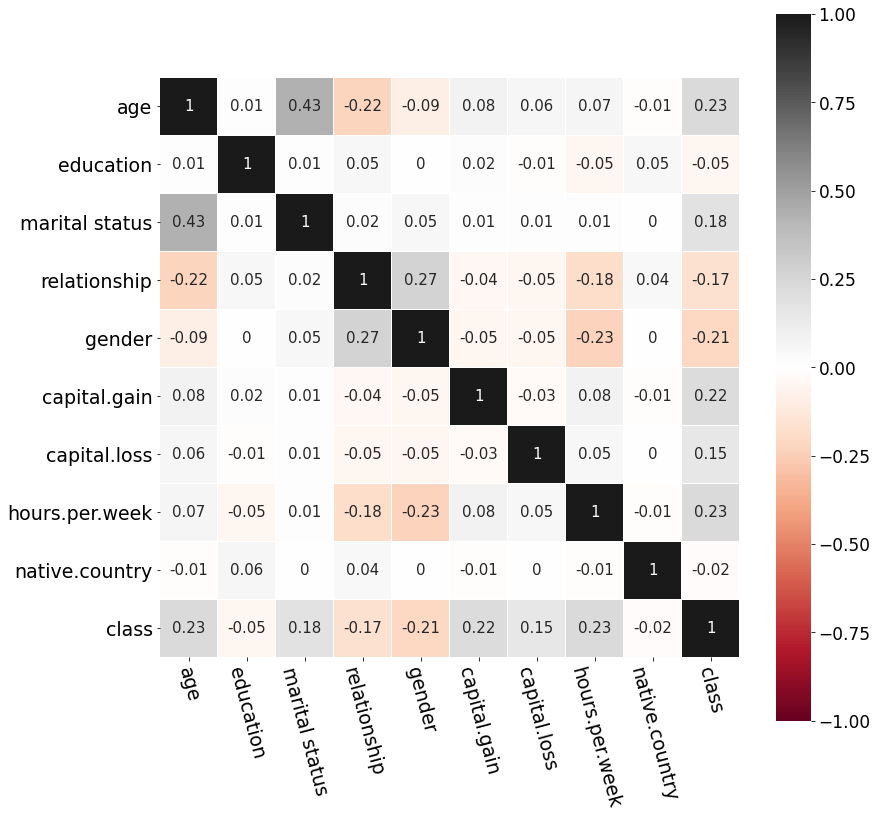}
	\centering
	\caption{Adult dataset: Pearson Correlation coefficients.}
	\label{fig:attCorr}
\end{figure}

In this section, we further investigate the accuracy-driven and fairness-oriented construction of FAHT and 2CFAHT. Figure~\ref{fig:attCorr} shows the Pearson correlation coefficients between each attribute as well as class label of the Adult dataset (we focus on the Adult dataset as the illustrating example for efficiency). Note that only attributes being used as splitting attributes during the constructions of HT as well as FAHT and 2CFAHT are shown, and correlation between each pair of variables is computed using all complete pairs of observations on those variables. In addition, different from static decision tree induction, an attribute will not be relaxed after being selected as the splitting attribute at one particular node and is a candidate splitting attribute for the succeeding splitting selection as well. 


In the constructed HT, ``capital-gain'' is selected as the root node and is re-selected several times at the succeeding nodes. This is quite intuitive as capital-gain is directly related with the amount of annual salary, i.e., class label. However, ``capital-gain'' also mirrors the intrinsic discrimination bias of the historic data as male, the favored community, has a higher likelihood of receiving higher capital-gain than female, the deprived community, based on the training data. On the other hand, FAHT selects ``age'' as the root node and re-selects it several times at the succeeding nodes as well. Generally speaking, income per annum increases as one gets older and more experienced, which holds regardless of the sensitive attribute value and from the correlation analysis. Selecting such an attribute balances encoding and diminishing discrimination of the training data for an accuracy-driven as well as fairness-oriented model. Such attribute selection strategy can also be concluded from Table~\ref{table: boundaryCorr_faht}, which shows the Pearson correlation coefficients between sensitive attributes and decision boundaries. As one can see, the sensitive attribute is more correlated with the predicted boundary of HT than FAHT's, because our model tries to build the boundary that the proportion of each community when receiving positive classification is identical or differs slightly, i.e., non-discriminatory. We also observe that the correlation between predicted boundary and actual boundary is stronger in HT than FAHT. This is because FAHT manages the trade-off between accuracy and discrimination when building the boundary while HT is completely accuracy-driven.

With respect to the attributes being selected for the construction of 2CFAHT, ``marital status'' is selected as its root attribute. Neither ``age'' nor ``marital status'' is discrimination-inclined compared to the intrinsic biased root attribute ``capital gain'' of HT. FAHT selected ``age'', generally speaking, is positively correlated with income per annum and holds that regardless of the sensitive attribute value. However, it is also possible that age could have local discrimination. That is to say, within a small age range, male could more likely to have a higher income than female as they tend to mature at different ages therefore differ in career age which could reflect income. 2CFAHT's learning idea of all groups being treated equally regardless of their population sizes to detect and reflect such type of discrimination encoding. Such fair attribute selection can also be concluded from the Pearson correlation coefficients between sensitive attribute and decision boundaries as shown in Table~\ref{table: boundaryCorr_2cfaht}. As one can see, the predicted boundaries of 2CFAHT are less correlated with the sensitive attribute than FAHT's as 2CFAHT also emphasizes local discrimination to maximize the cumulative fairness. In addition, different from the tree induction in static setting, the selected attributes are still splitting candidates for the succeeding splitting selection, such fairness-enhancing decisions therefore have impacts on following decisions as well (feedback loops) and could further enhance fairness-aware learning. We can also see that the decision boundary induced by 2CFAHT is more correlated with the actual boundary compared to FAHT's, as 2CFAHT assumes evolving distribution and adapts to the non-stationarity throughout the tree growth.     


%

\begin{table}[!htb]
	\centering
	\caption{Pearson Correlation coefficients between sensitive attribute, predicted decision boundary and actual decision boundary on Adult dataset. The values before colon are from the HT and after are from FAHT.}
	\label{table: boundaryCorr_faht}
	\begin{tabular}{cccc}
		\hline
		Entity & Sensitive attribute & Predicted boundary & Actual boundary\\
		\cline{1-4}
		Sensitive attribute & 1 : 1  &  -0.21 : -0.16 & -0.21 : -0.21 \\
		Predicted boundary &  -0.21 : -0.16  & 1 : 1 & 0.50 : 0.44\\
		Actual boundary & -0.21 : -0.21 & 0.50 : 0.44 & 1 : 1 \\
		\cline{1-4}
		\hline
	\end{tabular} 
\end{table}

\begin{table}[!htb]
	\centering
	\caption{Pearson Correlation coefficients between sensitive attribute, predicted decision boundary and actual decision boundary on Adult dataset. The values before colon are from the FAHT and after are from 2CFAHT.}
	\label{table: boundaryCorr_2cfaht}
	\begin{tabular}{cccc}
		\hline
		Entity & Sensitive attr& Predicted bndry & Actual bndry\\
		\cline{1-4}
		Sensitive attr & 1:1  &  -0.16 : -0.11  & -0.21 : -0.21 \\
		Predicted bndry &  -0.16 : -0.11  & 1 : 1 & 0.44 : 0.51\\
		Actual bndry & -0.21 : -0.21 & 0.44 : 0.51 & 1:1 \\
		\cline{1-4}
		\hline
	\end{tabular}
\end{table}

\begin{figure}[!htb]
	\centering
	\subfigure[Adult dataset]{
		\includegraphics[width=0.5\textwidth]{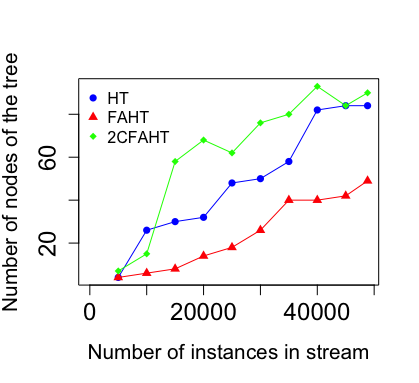}
	}%
	\subfigure[Census dataset]{
		\includegraphics[width=0.5\textwidth]{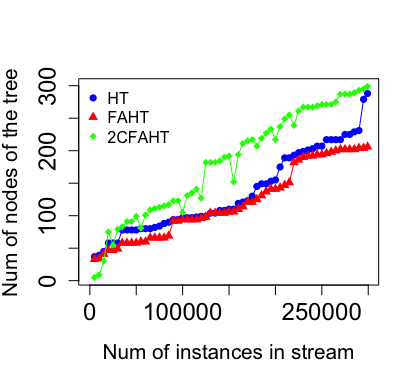}
	}
	\centering
	\caption{Model complexity (number of nodes) over the stream.}
	\label{fig:numOfNodes}
\end{figure}

What's more, theoretically, the induction of FAHT is expected to be more conservative than the normal HT. The reason is that the devised fair information gain splitting criterion takes class distribution and discrimination into consideration and is therefore more selective in evaluating candidate splitting attributes. As one can see from Figure~\ref{fig:numOfNodes}, FAHT results in a shorter tree comparing to HT, as its splitting criterion FIG is more restrictive comparing to IG. This conservative characteristic is also helpful for interpretation purpose. On the other hand, the model complexity of 2CFAHT fluctuates over the course of stream and is more complex than FAHT due to 2CFAHT's adaptation to stream. When the application scenario is stationary or prior stable distribution knowledge is available, FAHT therefore fits more than 2CFAHT as the latter's adaption incurs extra memory consumption.

\subsection{Analysis of Concept Adapting Components in 2CFAHT}
\label{sec: justification}

2CFHAT extends FAHT to process massive non-stationary data streams by introducing the adaptive fair information gain as well as adding concept drift detection and management mechanisms. To validate the theoretical design of these components, we incorporate AFIG into FAHT denoted as FAHT+ and 2CFAHT- representing 2CFAHT driven by FIG and compare them respectively. The results are shown in Table~\ref{table: criteriaComp}.

\begin{table}[!htb]
	\centering
	\caption{Accuracy-vs-discrimination between models assuming stationary and non-stationary based distributions. Percentage in parenthesis is the relative difference over the performance of its corresponding comparing method.}
	\label{table: criteriaComp}
	\begin{tabular}{ccccc}
		\hline
		\multirow{2}{*}{\diagbox{Methods}{Metric}}& \multicolumn{2}{c}{Adult dataset} & \multicolumn{2}{c}{Census dataset} \\ 
		\cline{2-5}
		&  Disc  & Acc & Disc & Acc \\
		\hline
		FAHT &  17.20\% & 81.62\%  & 3.63\% & 93.06\%\\
		\multirow{2}{*}{FAHT+}& 16.02\%  &  81.01\% & 2.61\% & 92.82\% \\ 
		&  (\textbf{-6.86\%})  & (\textbf{-0.75\%}) & (\textbf{-28.1\%}) & (\textbf{-0.26\%})   \\
		\hline
		\hline
		2CFAHT-  &19.14\%  & 83.76\%  & 2.20\% & 94.14\%  \\
		\multirow{2}{*}{2CFAHT}& 11.15\% &  85.01\% & 0.88\% & 95.03\% \\ 
		&  (\textbf{-41.75\%})  & (\textbf{+1.49\%}) & (\textbf{-60.0\%}) & (\textbf{+0.95\%}) \\
		\hline
	\end{tabular}
\end{table}

As shown in Table~\ref{table: criteriaComp}, it is clear that AFIG consistently enhances the fairness-aware learning in non-stationary online settings by effectively represent all types of discriminations while adapting to the evolving distribution over the course of stream, thus diminishing the discrimination to a lower level while maintaining a high prediction capability. The best discrimination reduction obtained by AFIG is 60.0\% on Census dataset. On the other hand, 2CFAHT's structural drift adaptation management also indeed pushes the discrimination to a lower level, which is consistent with its theoretical design. These enhanced anti-discrimination and concept-adapting abilities can also be statistically verified as shown in Table~\ref{table: McNemarstest_justification}.


\begin{table}[!htb]
	\centering
	\caption{The McNemar's test on the datasets for two different splitting criteria AFIG and FIG as well as two different tree growth strategies, which demonstrated the enhanced positive classification of the deprived group.}
	\label{table: McNemarstest_justification}
	\begin{threeparttable}[!htb]
		\begin{tabular}{ccccc}
			\hline
			\multirow{2}{*}{\diagbox{\textbf{FAHT}}{\textbf{FAHT+}}}& \multicolumn{2}{c}{{\textbf{Adult dataset}}$^1$} & \multicolumn{2}{c}{{\textbf{Census dataset}}$^2$}\\ 
			\cline{2-5}
			& Granted &  Rejected & Granted & Rejected\\
			\hline
			Granted &  716 & 110 & 1,120 & 263 \\
			Rejected &   173 & 15,193 & 468 & 153,924 \\
			\hline
		\end{tabular}
		\begin{tablenotes} 
			\item[1] Chi-squared = 13.583, df = 1, p-value = 0.0002282
			\item[2] Chi-squared = 56.93, df = 1, p-value $\textless$ 4.516e-14\\
		\end{tablenotes}
	\end{threeparttable}
	\centering
	
	\begin{threeparttable}[!htb]
		\begin{tabular}{ccccc}
			\hline
			\multirow{2}{*}{\diagbox{\textbf{2CFAHT-}}{\textbf{2CFAHT}}}& \multicolumn{2}{c}{{\textbf{Adult dataset}}$^3$} & \multicolumn{2}{c|}{{\textbf{Census dataset}}$^4$}\\ 
			\cline{2-5}
			& Granted &  Rejected & Granted & Rejected\\
			\hline
			Granted &  1,127 & 80 & 1,331 & 359  \\
			\hline
			Rejected &  153 & 14,832 & 658 & 153,427 \\
			\hline
		\end{tabular}
		\begin{tablenotes} 
			\item[3] Chi-squared = 22.249, df = 1, p-value = 2.395e-06
			\item[4] Chi-squared = 87.32, df = 1, p-value $\textless$ 2.2e-16
		\end{tablenotes}
	\end{threeparttable}

\end{table}


\begin{figure}[ht]
	\centering
		\subfigure[Auldt dataset]{
			\includegraphics[width=0.4\textwidth]{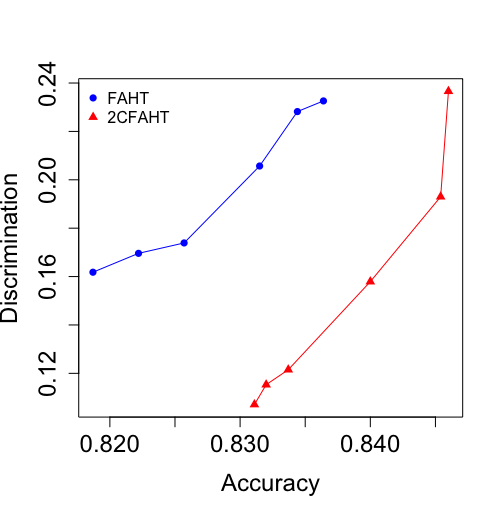}
		}%
		\subfigure[Census dataset]{
			\includegraphics[width=0.4\textwidth]{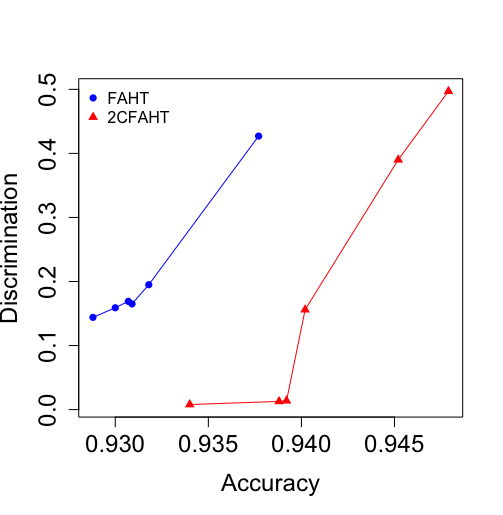}
		}
		\centering
		\caption{The accuracy and discrimination trade-off observed by adjusting the tunable parameter $\gamma$ ranging from 100000 to 1 with step size 10.}
		\label{fig: custom}
	\end{figure}

\subsection{Fine-grained Fairness Control}

We also experiment with 2CFAHT's fine-grained control of the trade-off between fairness and accuracy for the sake of instantiating application-wise fairness-aware learning. We adjust then fix the value of $\gamma$ to manage the trade-off between fairness and accuracy. That is to say, the value of $\gamma$ is predefined and does not adapt to reflect the prior learning focus to meet the current performance constraint. The AFIG is also incorporated into FAHT for an additional validation. Figure~\ref{fig: custom} visualizes the results by showing the fairness and accuracy subject to fairness constraints with different values of $\gamma$. As expected, as the value of $\gamma$ decreases, i.e., the current application focuses more on accuracy, the accuracy increase is accompanied by the decrease in fairness.

\section{Discussion and limitation}

While this work focuses on binary classification with a single sensitive attribute, this setting reflects the dominant formulation in fairness-aware online learning due to its conceptual clarity and computational tractability. However, many real-world applications involve multi-class classification tasks and multiple sensitive attributes, which introduce additional complexity. In particular, extending fairness-aware split criteria to multi-class settings requires redefining fairness notions (e.g., multi-class extensions of statistical parity or equal opportunity) and adapting evaluation metrics accordingly. Similarly, handling multiple sensitive attributes may require decomposing or aggregating group-based disparities in more nuanced ways. These extensions represent important but distinct research directions. Future work could explore generalizing the fair information gain and adaptive splitting mechanisms to support structured output spaces and intersectional fairness across multiple attributes, potentially leveraging tensor-based representations or fairness-aware multi-objective optimization.

In addition, while our framework focuses on statistical parity, it is designed to flexibly accommodate alternative group fairness notions such as equal opportunity or disparate impact by adjusting the fairness gain component in the splitting criteria. However, different fairness metrics are not always aligned; optimizing for one may degrade performance on another. For example, statistical parity aims to equalize outcomes across groups, while equal opportunity focuses on equalizing true positive rates. In streaming contexts, selecting the most appropriate metric often depends on domain-specific goals and constraints. Our use of statistical parity serves as a representative baseline, but future work could explore adaptive or application-specific fairness selection mechanisms. We note that individual fairness, which relies on instance-level similarity, poses additional challenges in online settings and remains an open research direction.

Another important consideration is the computational cost associated with maintaining fairness-aware structures in streaming environments. The 2CFAHT model, in particular, introduces surrogate subtrees and adaptive $\gamma$ updates to support fairness under concept drift. While this design enables fine-grained fairness control, it introduces additional memory and computational overhead compared to baseline learners. This includes the cost of maintaining parallel structures, tracking fairness metrics over time, and dynamically adjusting model behavior in response to fairness-performance trade-offs. Our empirical results in Section 6.3 show that the method remains practical for real-world streams. However, in high-throughput or resource-constrained settings—such as edge devices or embedded platforms—further optimization is needed. Future work could explore lightweight drift detectors such as ADWIN, Page-Hinkley, or DDM, which monitor performance changes efficiently with minimal memory. In place of persistent surrogate subtrees, pruning and regrowing branches upon confirmed drift may offer a more scalable alternative. Compact representations of fairness statistics, such as histograms, sketches, or sliding windows, can further reduce memory usage. These optimizations will be essential for improving the deployability of fairness-aware streaming models without compromising adaptability or fairness guarantees.

While our framework is designed to adapt to evolving data distributions via dynamic $\gamma$ adjustment, it does not explicitly distinguish among different types of concept drift, such as sudden, gradual, incremental, or recurring drift. Each of these drift types may have different implications for fairness dynamics and model stability. Integrating fine-grained drift classification techniques could enable more tailored responses, allowing the model to calibrate its adaptation strategy depending on the nature and severity of the distributional shift. For instance, sudden drift may call for more aggressive $\gamma$ updates, while gradual drift may benefit from smoother transitions. Furthermore, understanding the temporal behavior of the self-adapting $\gamma$ is important for assessing the stability and reliability of fairness adaptation. Frequent or erratic changes in $\gamma$ could lead to unstable predictions or overreaction to short-term fluctuations. Incorporating stability diagnostics—such as monitoring the variance of $\gamma$ over time or correlating its evolution with fairness-accuracy trade-offs—would improve transparency and trustworthiness. These extensions, though beyond the current scope, represent promising directions for future work to strengthen the robustness and interpretability of fairness-aware streaming decision trees.

Interpretability, while commonly associated with decision trees, is not the primary focus of this work. Our emphasis lies in achieving fairness-aware learning under streaming and non-stationary conditions. Nonetheless, we recognize that the increased structural complexity of 2CFAHT may reduce transparency in certain use cases. Future research could explore incorporating interpretability metrics where needed, especially for deployment in regulated or high-stakes domains.

With respect to high-dimensional settings, 2CFAHT benefits from the scalability properties of Hoeffding Trees, which support efficient attribute evaluation in streaming environments. However, in scenarios with many noisy or weakly predictive features, additional strategies may be needed. One promising direction is to combine fairness-aware trees with online feature selection or dimensionality reduction to maintain efficiency and model quality in complex data streams.

As with many fairness-aware learning frameworks, our method assumes that the incoming data stream is representative of the underlying population and not influenced by prior model decisions. However, in many real-world applications—such as credit scoring, content recommendation, and predictive policing—the model's outputs can affect user behavior or system dynamics, thereby altering future input distributions. This creates the potential for feedback loops that reinforce biases or amplify disparities over time, even when fairness constraints are applied at each step. For example, an applicant denied a loan due to predicted risk may not reapply, shifting the distribution of future training data and possibly entrenching unfair patterns. Our current framework does not explicitly model or correct for such feedback effects, and we acknowledge this as a key limitation. Addressing feedback loops in streaming environments requires modeling the interaction between decision policies and data generation, which may necessitate causal reasoning, dynamic system modeling, or reinforcement learning techniques. Recent advances in fair reinforcement learning and adaptive causal inference offer promising foundations for mitigating these challenges. We view the integration of fairness-aware decision-making with causal and policy-aware models as a crucial next step toward robust, real-world deployability.

\section{Conclusion}
\label{sec: conclusion}

This paper focuses on the highly under-explored discrimination-aware learning in massive data streams. To address the unique challenges, we propose the first online version of fair decision trees with constraints to fulfill different online fair decision-making requirements. Our design also includes a mechanism for adjusting the jointly learning focus on the fly depending on the respective non-stationary implications on prediction accuracy and fairness performance, thus be robust on the evolving implications of data encoding and bias reduction. The positive results of conducted experiments show the flexibility and versatility of FAHT and 2CFAHT in massive discriminated data stream settings. Immediate future directions include to extend these results for the unbiased scene graph generation as well as the more challenging unsupervised fair clustering domain.

\section{Conflict of Interest}

The authors affirm that they do not have any known financial interests or personal relationships that could have potentially affected the integrity of the work presented in this paper.


%
%

\nocite{*}
\bibliographystyle{spmpsci}      
\bibliography{typeinst}   


\end{document}